 \documentclass[tablecaption=bottom,wcp]{jmlr} % W&CP article

\usepackage{booktabs}
\usepackage[load-configurations=version-1]{siunitx} % newer version
\usepackage[scaled=.86]{helvet} % setting sans serif font to {Helvetica}, scaled to proper x-height
\usepackage{amsmath} % enabling various math envs
\DeclareMathOperator{\const}{const}
\newcommand\diff{\mathop{}\!\mathrm{d}}

\usepackage{bm}
\usepackage[noamssymbols]{newtxmath} % various optional arguments
\usepackage[cal=euler,calscaled=1.0,scr=boondoxo,scrscaled=1.0]{mathalfa}
\usepackage{makecell} % enabling fancy commands for table cells
\usepackage{xcolor} % enabling more colors for comments
\theorembodyfont{\upshape}
\theoremheaderfont{\scshape}
\theorempostheader{:}
\theoremsep{\newline}

\jmlrproceedings{AABI 2018}{1st Symposium on Advances in Approximate Bayesian Inference, 2018}

\title[Parallel-tempered Stochastic Gradient HMC]{Parallel-tempered Stochastic Gradient Hamiltonian Monte Carlo for Approximate Multimodal Posterior Sampling}

 \author{\Name{Rui Luo\nametag{\thanks{Equal}}} \Email{rui.luo@aig.com}\\
   \Name{Qiang Zhang\nametag{\footnotemark[1]}} \Email{qiang.zhang@aig.com}\\
   \Name{Yuanyuan Liu} \Email{yuanyuan.liu@aig.com}\\
   \addr American International Group, Inc.}

\begin{document}

\maketitle

\vspace{-1.5em}
\begin{abstract}
We propose a new sampler that integrates the protocol of parallel tempering with the Nos\'e-Hoover (NH) dynamics. The proposed method can efficiently draw representative samples from complex posterior distributions with multiple isolated modes in the presence of noise arising from stochastic gradient. It potentially facilitates deep Bayesian learning on large datasets where complex multimodal posteriors and mini-batch gradient are encountered.
\end{abstract}

% Keywords may be removed
%\begin{keywords}
%List of keywords
%\end{keywords}

\section{Introduction}
\label{sec:intro}

In Bayesian inference, one of the fundamental problems is to efficiently draw \emph{i.i.d.} samples from the posterior distribution $\pi(\theta | \mathscr{D})$ given the dataset $\mathscr{D} = \{x\}$, where $\theta \in \mathbb{R}^D$ denotes the variable of interest. Provided the prior distribution $\pi(\theta)$ and the likelihood per datum $\ell(\theta; x)$, the posterior to be sampled can be formulated as
\vskip -0.3in
\begin{align}
\pi(\theta | \mathscr{D}) = \pi(\theta) \prod_{x \in \mathscr{D}} \ell(\theta; x).%~\mbox{with the likelihood per datum}~\ell(\theta; x).
\end{align}
\vskip -0.1in\noindent

To facilitate posterior sampling, the framework of Markov chain Monte Carlo (MCMC) has been established, which has initiated a broad family of methods that generate Markov chains to propose new sample candidates and then apply tests of acceptance in order to guarantee the condition of detailed balance. Methods like the Metropolis-Hastings (MH) algorithm \citep{metropolis1953equation,hastings1970monte}, the Gibbs sampler \citep{geman1984stochastic}, and the hybrid/Hamiltonian Monte Carlo (HMC) \citep{duane1987hybrid,neal2011mcmc} are famous representatives for the MCMC family where different generating procedures of Markov chains are adopted; each of those methods has achieved great success on various tasks in statistics and related fields.

Among MCMC methods, HMC, in particular, has attracted attention due to its exploitation of gradient information. In a typical HMC setting \citep{neal2011mcmc}, the target posterior distribution $\pi(\theta | \mathscr{D})$ is embedded into a virtual physical system fixed at the standard temperature $T = 1$ with the potential energy defined in the form of
\vskip -0.275in
\begin{align}
\label{eq:U}
U(\theta) = -\log \pi(\theta | \mathscr{D}) = -\log \pi(\theta) - \sum_{x \in \mathscr{D}} \log \ell(\theta; x) - \const.
\end{align}
\vskip -0.1in\noindent
The variable of interest $\theta$ is interpreted as the position of the system in the phase space; an auxiliary variable $p \in \mathbb{R}^D$ is then introduced as the conjugate momentum corresponding to the kinetic energy $p^{\top}M^{-1}p/2$. By defining the total energy, i.e. the Hamiltonian, as the sum of the potential and kinetic energy, the Hamiltonian dynamics that governs the physical system can therefore be derived from the Hamilton's formalism. From the perspective of sampling, new sample candidates are proposed via simulating the Hamiltonian dynamics, where the gradient of potential $\nabla U(\theta)$ is utilized.

Despite possessing numerous advantages against its alternatives within the MCMC family, HMC still suffers, however, from two major issues: 1. gradient noise arising from mini-batches may lead to a severe deviation of the dynamics from the desired orbit; 2. isolated modes may not be correctly sampled or even left undiscovered. Unfortunately, as one deals with deep neural networks training on large datasets, those two problems arise simultaneously: deep neural networks leads to complex posterior distributions for the parameters, which may contain numbers of isolated modes; efficient training on large datasets requires mini-batching, the gradient hence would be quite noisy as is evaluated on a small fraction of dataset.

It has long been known that the tempering mechanism is capable of helping the system to get across high energy barriers and hence improve the ergodicity \citep{marinari1992simulated,earl2005parallel}. Recently, the research of incorporating tempering into MCMC methods has provided a practical approach towards efficient multimodal posterior sampling \citep{DBLP:conf/uai/GrahamS17,luo2018thermostat}. In the meantime, the advances in thermostatting techniques for molecular dynamics \citep{jones2011adaptive} have shed some light on adaptive control for noisy dynamics. In this paper, we propose a novel method that addresses the two issues previously mentioned for HMC; it combines the protocol of parallel tempering \citep{swendsen1986replica,sugita1999replica} with the dynamics of Nos\'e-Hoover (NH) thermostat \citep{nose1984unified,hoover1985canonical}. The simulation shows the advantages w.r.t. the accuracy as well as efficiency of our method against the classic HMC \citep{neal2011mcmc} and one of its stochastic variants, Stochastic Gradient Nos\'e-Hoover Thermostat (SGNHT) \citep{ding2014bayesian}.

%%%%%%%%%%%%%%%%%%%%

\vspace{-0.25em}
\section{Parallel-tempered Stochastic Gradient Hamiltonian Monte Carlo}
\label{sec:mReMC}

The proposed method consists of two alternating subroutines: 1. the parallel dynamics simulation of system replicas, and 2. the configuration exchange between replicas. The first subroutine utilizes the Nos\'e-Hoover thermostat to adaptively detect and neutralize the noise within mini-batch gradient; the second incorporates a mini-batch acceptance test to ensure the detailed balance during exchanges.

\vspace{-0.25em}
\subsection{Parallel Dynamics Simulation of System Replicas}

We define an increasing ladder $\{T_j\}_{j=1}^R$ of temperature with $R$ rungs; the temperature ranges from the standard $T_1 = 1$ to some higher temperature. On each rung $j$, a replica $(\theta_j, p_j)$ of the physical system is initialized and the actual potential energy for that replica is rescaled to $U(\theta_j)/T_j$.

As the datum $x$ within each mini-batch $\mathscr{S}$ is independently selected at random, the mini-batch gradient can be approximated by a Gaussian variable due to the Central Limit Theorem (CLT):
\vskip -0.225in
\begin{align}
\nabla\tilde{U}(\theta) &= -\nabla\log \pi(\theta) - \frac{|\mathscr{D}|}{|\mathscr{S}|}\sum_{x \in \mathscr{S} \subset \mathscr{D}} \nabla\log \ell(\theta; x).
\end{align}
\vskip -0.1in\noindent

To retain the correct trajectory in simulating the system dynamics, we leverage the NH thermostat because of its capability of adaptive control of the gradient noise \citep{jones2011adaptive,ding2014bayesian}. According to the formulation of \cite{hoover1985canonical}, for each replica $(\theta_j, p_j)$, we augment the system with NH thermostat $\xi_j \in \mathbb{R}$ and then modify the dynamics as:
{
\vskip -0.20in
\begin{align}
\label{eq:nht}
\frac{\diff{\theta_j}}{\diff{t}} = M^{-1} p_j,~~~~\frac{\diff{p_j}}{\diff{t}} = -\nabla\tilde{U}(\theta_j)/T_j - \xi p_j,~~~~\frac{\diff{\xi_j}}{\diff{t}} = \left. \left[ p_j^{\top} M^{-1} p_j - D \right] \middle/ Q, \right.
\end{align}
%\vskip -0.1in\noindent
}%
where $M$ denotes the mass, and $Q$ the thermal inertia. It can be proved that the dynamics in Eq. \eqref{eq:nht} leads to a stationary distribution w.r.t. $\theta_j$ by the Fokker-Planck equation \citep{risken1989fpe} 
\vskip -0.225in
\begin{align}
\pi_j(\theta_j) \propto e^{-U(\theta_j)/T_j}.
\end{align}
\vskip -0.05in\noindent
This guarantees that, during the simulation, one can readily recover the desired distribution at a certain temperature $T_j$ by simply retaining the position $\theta_j$ and discarding the momentum $p_j$ as well as the thermostat $\xi_j$. Note that for the replica on rung $1$, the temperature is fixed at standard $T_1 = 1$ and the position $\theta_1 = \theta$ is distributed as the target posterior $\pi_1(\theta_1) = e^{-U(\theta_1)/T_1} = e^{-U(\theta)} = \pi(\theta | \mathscr{D})$.

\vspace{-0.25em}
\subsection{Configuration Exchange between Replicas}

The principles of statistical physics suggest that high temperature facilitates the physical systems to get across energy barriers, which means replicas at higher temperatures are more likely to traverse among different modes of the distributions. As a consequence, however, the distribution sampled at high temperature has a spread spectrum and is hence biased. To recover an unbiased distribution, we perform configuration exchange between replicas at higher temperatures and the one at the standard.

Consider the configuration exchange between the replicas on rung $i$ and $j$; as is a non-physical process, the exchange has to satisfy the condition of detailed balance:
\vskip -0.225in
\begin{align}
\label{eq:balance}
\pi_j(\theta_j)\pi_k(\theta_k)\alpha[(j, k) \to (k, j)] = \pi_j(\theta_k)\pi_k(\theta_j)\alpha[(k, j) \to (j, k)],
\end{align}
\vskip -0.05in\noindent
where the transition probability reads
\vskip -0.225in
\begin{align}
\alpha[(i, j) \to (j, i)] = \frac{\pi_j(\theta_k)\pi_k(\theta_j)}{\pi_j(\theta_j)\pi_k(\theta_k) + \pi_j(\theta_k)\pi_k(\theta_j)} = \frac{1}{1 + e^{-\delta E}},
\end{align}
\vskip -0.05in\noindent
and $\delta E = \big[U(\theta_k) - U(\theta_j)\big]\big[(T_k - T_j) / T_jT_k\big]$. It is straightforward to verify that Eq. \eqref{eq:balance} holds. Note that the transition probability $\alpha[(j, k) \to (k, j)]$ resembles the logistic distribution; such logistic test of acceptance is developed by \cite{barker1965monte}.

With mini-batching, the potential energy $\tilde{U}(\theta_j)$ becomes a r.v., and so is the difference $\tilde{U}(\theta_k) - \tilde{U}(\theta_j)$. By CLT, $\delta E$ is asymptotically Gaussian with some certain variance $\sigma^2$. \cite{DBLP:conf/uai/SeitaPCC17} proposed a mini-batch version of Baker's logistic test of acceptance such that $\delta E + \mathscr{C} > 0$ must hold for the exchange to carry out, where $\mathscr{L}$ denotes an auxiliary correction r.v. that aims to bridge the gap between the logistic distribution and Gaussian. The probability density $p_{\mathscr{C}}$ of this correction variable $\mathscr{C}$ satisfies the convolution equation $p_{\mathscr{C}} * p_{\mathscr{N}_{\sigma^2}} = p_{\mathscr{L}}$; it is equivalent to solve the Gaussian deconvolution problem w.r.t. the standard logistic distribution.

With the convolution theorem for distributions, it is helpful to convert the Gaussian deconvolution into solving for the inverse Fourier transform w.r.t. quotient of characteristic functions
\vskip -0.225in
\begin{align}
\label{eq:exact}
p_{\mathscr{C}} = \frac{1}{2\pi} \int_{-\infty}^{\infty} \frac{\phi_{\mathscr{L}}(t)}{\phi_{\mathscr{N}_{\sigma^2}}(t)}e^{-ixt}\diff{t},~\quad\mbox{since}~~\phi_{\mathscr{C}} = \phi_{\mathscr{L}} / \phi_{\mathscr{N}_{\sigma^2}},
\end{align}
\vskip -0.10in\noindent
where $\phi_{\mathscr{N}_{\sigma^2}}$ and $\phi_{\mathscr{L}}$ denote the characteristic functions of $\mathscr{N}(0, \sigma^2)$ and the standard logistic r.v., respectively. As the logistic distribution has much heavier tails than the Gaussian, the exact solution of $p_{\mathscr{C}}$ does not exist: the ``integrand'' on the RHS of Eq. \eqref{eq:exact} is in fact not integrable. We can only approximate $p_{\mathscr{C}}$ by introducing the kernel $\psi = e^{-\gamma^2 t^4}$ of bandwidth $1/\gamma$ \citep[see][]{fan1991optimal} in Eq. \eqref{eq:exact}:
\vskip -0.225in
\begin{align}
\label{eq:approx}
\hat{p}_{\mathscr{C}} = \frac{1}{2\pi} \int_{-\infty}^{\infty} \psi \cdot \frac{\phi_{\mathscr{L}}}{\phi_{\mathscr{N}_{\sigma^2}}} e^{-itx} \diff{t} = \frac{1}{2\pi} \int_{-\infty}^{\infty} \bigg[ \frac{\psi}{\phi_{\mathscr{N}_{\sigma^2}}} \bigg] \phi_{\mathscr{L}} e^{-ixt} \diff{t}.
\end{align}
\vskip -0.10in\noindent

Using the Hermite polynomials $H_k$ \citep{abramowitz1965handbook}, we now expand the quotient within the brackets of Eq. \eqref{eq:approx} as
%\vskip -0.175in
\begin{align}
\frac{\psi}{\phi_{\mathscr{N}_{\sigma^2}}} = e^{-\gamma^2 t^4 + \sigma^2 t^2 / 2} = \sum_{k = 0}^{\infty} \frac{\gamma^k}{k!}H_k(\sigma^2 / 4\gamma) t^{2k}.
\end{align}
%\vskip -0.10in\noindent
The correction distribution can be approximated via Fourier's differential theorem:
{\small
%\vskip -0.175in
\begin{align}
\hat{p}_{\mathscr{C}} = \sum_{k = 0}^{\infty} \frac{(-1)^k}{k!} H_k(\sigma^2 / 4\gamma) \gamma^k \left[ \frac{1}{2\pi} \int_{-\infty}^{\infty} (-it)^{2k} \phi_{\mathscr{L}} e^{-itx} \diff{t} \right] = \sum_{k = 0}^{\infty} \frac{(-1)^k}{k!} H_k(\sigma^2/4\gamma) \gamma^k p_{\mathscr{L}}^{(2k)},
\end{align}
%\vskip -0.10in\noindent
}%
where $p_{\mathscr{L}}^{(j)}$ represents the $(j+1)$-th derivative of logistic function, which can be efficiently calculated in a recursive fashion \citep{minai1993derivatives}.

%\begin{align}
%\xi_{\mathscr{L}} = \xi_{\mathscr{N}} + \xi_{\mathscr{C}}\\
%p_{\mathscr{L}} = p_{\mathscr{N}} * p_{\mathscr{C}}\\
%\phi_{\mathscr{L}} = \phi_{\mathscr{N}} \cdot \phi_{\mathscr{C}}\\
%\phi_{\mathscr{C}} = \phi_{\mathscr{L}} / \phi_{\mathscr{N}}\\
%\phi_{\mathscr{L}} = \frac{\pi t}{\sinh{\pi t}}\\
%\phi_{\mathscr{N}} = e^{-\sigma^2 t^2 / 2}\\
%\hat{\phi}_{\mathscr{C}} = \psi \cdot \phi_{\mathscr{L}} / \phi_{\mathscr{N}}\\
%\psi = e^{-\gamma^2 t^4}\\
%\hat{p}_{\mathscr{C}} = \frac{1}{2\pi} \int_{-\infty}^{\infty} \hat{\psi}_{\mathscr{C}} e^{-itx} \diff{t}\\
%= \frac{1}{2\pi} \int_{-\infty}^{\infty} \frac{\psi}{\phi_{\mathscr{N}}} \phi_{\mathscr{L}} e^{-itx} \diff{t}\\
%\frac{\psi}{\phi_{\mathscr{N}}} = e^{-\gamma^2 t^4 + \sigma^2 t^2 / 2} = \sum_{k = 0}^{\infty} \frac{H_k(\frac{\sigma^2}{4\gamma})}{k!}\gamma^k t^{2k}\\
%e^{-u^2 + 2xu} = \sum_{k = 0}^{\infty} \frac{H_k(x)}{k!} u^k\\
%\hat{p}_{\mathscr{C}} = \frac{1}{2\pi} \int_{-\infty}^{\infty} \left[ \frac{\psi}{\phi_{\mathscr{N}}} \right] \phi_{\mathscr{L}} e^{-itx} \diff{t}\\
%= \frac{1}{2\pi} \int_{-\infty}^{\infty} \left[ \sum_{k = 0}^{\infty} \frac{H_k(\frac{\sigma^2}{4\gamma})}{k!} \gamma^k t^{2k} \right] \phi_{\mathscr{L}} e^{-itx} \diff{t}\\
%= \sum_{k = 0}^{\infty} \frac{(-1)^k}{k!} H_k\left(\frac{\sigma^2}{4\gamma}\right) \gamma^k \left[ \frac{1}{2\pi} \int_{-\infty}^{\infty} (-it)^{2k} \phi_{\mathscr{L}} e^{-itx} \diff{t} \right]\\
% = \sum_{k = 0}^{\infty} \frac{(-1)^k}{k!} H_k\left(\frac{\sigma^2}{4\gamma}\right) \gamma^k p_{\mathscr{L}}^{(2k)}
%\end{align}

\begin{figure}[h]
\centering
\includegraphics[width=0.625\columnwidth]{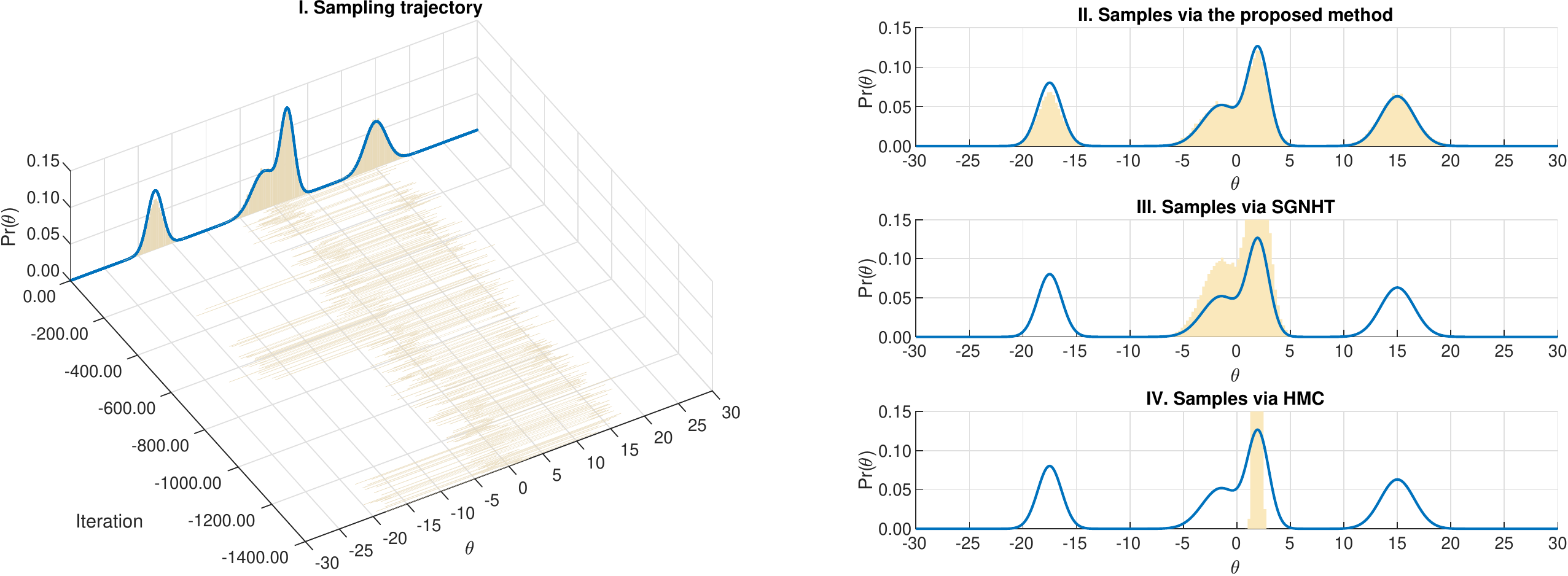}
\caption{Experiment on sampling a $1d$ mixture of 4 Gaussians.}
\label{fig:1d}
\vspace{-20pt}
\end{figure}

\begin{figure}[h]
\centering
\includegraphics[width=0.775\columnwidth,height=7em]{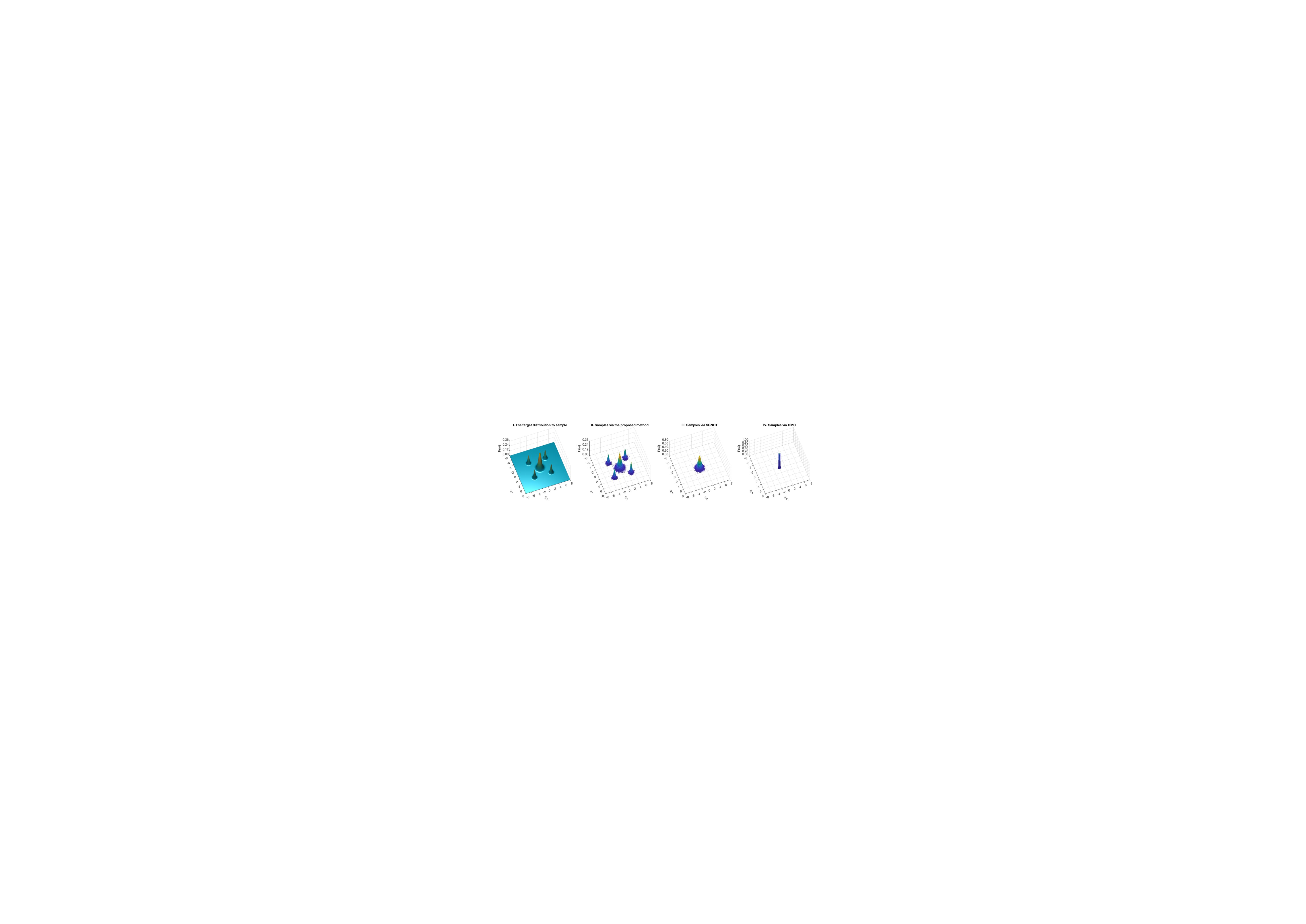}
\caption{Experiment on sampling a $2d$ mixture of 5 Gaussians.}
\label{fig:2d}
\vspace{-20pt}
\end{figure}

\section{Experiment}
\label{sec:experiment}

We conduct two sets of experiments on synthetic distributions: the first is a mixture of $4$ Gaussians in $1d$, and the second is a $2d$ Gaussian mixture with $5$ isolated modes. The potential energy as well as its gradient is perturbed by zero-mean Gaussian noise with variance $\sigma^2 = 0.25$, and all samplers in test have no access to the actual parameters of that noise. We establish a ladder of temperature with $R = 10$ rungs ranging from $T_1 = 1$ to $T_R = 10$, i.e. totally $10$ replicas are simulated in parallel. The baselines are the classic HMC \citep{neal2011mcmc} the adaptive variant SGNHT \citep{ding2014bayesian}. It is demonstrated in Fig. \ref{fig:1d} and \ref{fig:2d} that, in both synthetic testing cases, our method has accurately sampled the target distributions with multiple isolated modes in the presence of noise within mini-batch gradient, where all baselines failed: SGNHT managed to control the gradient noise but did not discover the isolated modes while the classic HMC appears to be unable to correctly draw samples due to the deviated dynamics. Moreover, the subplot on the \emph{left} of Fig. \ref{fig:1d} illustrates the sampling trajectory of our method, indicating a good mixing property.

\bibliography{aabi_2018}

\end{document}